# Hypergraph and protein function prediction with gene expression data

Loc Tran

University of Minnesota

Abstract: Most network-based protein (or gene) function prediction methods are based on the assumption that the labels of two adjacent proteins in the network are likely to be the same. However, assuming the pairwise relationship between proteins or genes is not complete, the information a group of genes that show very similar patterns of expression and tend to have similar functions (i.e. the functional modules) is missed. The natural way overcoming the information loss of the above assumption is to represent the gene expression data as the hypergraph. Thus, in this paper, the three un-normalized, random walk, and symmetric normalized hypergraph Laplacian based semi-supervised learning methods applied to hypergraph constructed from the gene expression data in order to predict the functions of yeast proteins are introduced. Experiment results show that the average accuracy performance measures of these three hypergraph Laplacian based semi-supervised learning methods are the same. However, their average accuracy performance measures of these three methods are much greater than the average accuracy performance measures of un-normalized graph Laplacian based semi-supervised learning method (i.e. the baseline method of this paper) applied to gene co-expression network created from the gene expression data.

## I. Introduction

Protein function prediction plays a very important role in modern biology. Detecting the function of proteins by biological experiments is very time-consuming and difficult. Hence a lot of computational methods have been proposed to infer the functions of the proteins by using various types of information such as gene expression data and protein-protein interaction networks [1].

The classical way predicting protein function infers the similarity to function from sequence homologies among proteins in the databases using sequence similarity algorithms such as FASTA [2] and PSI-BLAST [3]. Next, to predict protein function, the natural model of relationship between proteins or genes which is graph can also be employed. This model can be protein-protein interaction network or gene co-expression network. In this model, the nodes represent proteins or genes and the edges represent for the possible interactions between nodes. Then, machine learning methods such as Support Vector Machine [5], Artificial Neural Networks [4], un-normalized graph Laplacian based semi-supervised learning method [6], the symmetric normalized and random walk graph Laplacian based semi-supervised learning methods [7], or neighbor counting method [8] can be applied to this graph to infer the functions of un-annotated protein. While the neighbor counting method labels the protein with the function that occurs frequently in the protein's adjacent nodes in the protein-protein interaction network and hence does not utilized the full topology of the network, the Artificial Neural Networks, Support Vector Machine, un-normalized, symmetric normalized and random walk graph Laplacian based semi-supervised learning method utilizes the full topology of the network and the Artificial Neural Networks and Support Vector Machine are supervised learning methods. While the neighbor counting method, the Artificial Neural Networks, and the three graph Laplacian based semi-supervised learning methods are all based on the assumption that the labels of two adjacent proteins in graph are likely to be the same, SVM do not rely on this assumption. Unlike graphs used in neighbor counting method, Artificial Neural Networks, and the three graph Laplacian based semi-supervised learning methods are very sparse, the graph (i.e. kernel) used in SVM is fully-connected.

While Artificial Neural Networks method is applied to the single protein-protein interaction network, the SVM method and three graph Laplacian based semi-supervised learning methods try to use weighted combination of multiple networks (i.e. kernels) such as gene co-expression network and protein-protein interaction network to improve the accuracy performance measures. While [5] (SVM method) determines the optimal weighted combination of networks by solving the semi-definite problem and [6] (un-normalized graph Laplacian based semi-supervised learning method) uses a dual problem and gradient descent to determine the weighted combination of networks, [7] uses the integrated network combined with equal weights, i.e. without optimization due to the integrated network combined with optimized weights has similar performance to the integrated network combined with equal weights and the high time complexity of optimization methods.

The un-normalized, symmetric normalized, and random walk graph Laplacian based semi-supervised learning methods are developed based on the assumption that the labels of two adjacent proteins or genes in the network are likely to be the same [6]. In this paper, we use gene expression data for protein function prediction problem. Hence this assumption can be interpreted as pairs of genes showing a similar pattern of expression and thus sharing edges in a gene co-expression network tend to have similar function. However, assuming the pairwise relationship between proteins or genes is not complete, the information a group of genes that show very similar patterns of expression and tend to have similar functions [8] (i.e. the functional modules) is missed. The natural way overcoming the information loss of the above assumption is to represent the gene expression data as the hypergraph [9, 10]. A hypergraph is a graph in which an edge (i.e. a hyper-edge) can connect more than two vertices. In [9, 10], the symmetric normalized hypergraph Laplacian based semi-supervised learning method have been developed and successfully applied to text categorization and letter recognition applications. To the best of my knowledge, the hypergraph Laplacian based semi-supervised learning methods have not yet been applied to protein function prediction problem. In this paper, we will develop the symmetric normalized, random walk, and un-normalized hypergraph Laplacian based semi-supervised learning methods and apply these three methods to the hypergraph constructed from gene expression data available from [11] by applying k-mean clustering method to this gene expression dataset.

We will organize the paper as follows: Section 2 will introduce the definition hypergraph Laplacians and their properties. Section 3 will introduce the un-normalized, random walk, and symmetric normalized hypergraph Laplacian based semi-supervised learning algorithms in detail. Section 4 will show how to derive the closed form solutions of normalized and un-normalized hypergraph Laplacian based semi-supervised learning algorithm from regularization framework. In section 5, we will apply the un-normalized graph Laplacian based semi-supervised learning algorithm (i.e. the current state of art method applied to protein function prediction problem) to gene co-expression network created from gene expression data available from [11] and compare its accuracy performance measure to the three hypergraph Laplacian based semi-supervised learning algorithms' accuracy performance measures. Section 6 will conclude this paper and the future direction of researches of this protein function prediction problem utilizing discrete operator of graph will be discussed.

## II. Hypergraph definitions

Given a hypergraph G=(V,E), where V is the set of vertices and E is the set of hyper-edges. Each hyper-edge $e \in E$ is the subset of V. Please note that the cardinality of e is greater than or equal two. In the other words, $|e| \geq 2$, for every $e \in E$. Let w(e) be the weight of the hyper-edge e. Then W will be the $R^{|E|*|E|}$ diagonal matrix containing the weights of all hyper-edges in its diagonal entries.

### 2.1 Definition of incidence matrix H of G

The incidence matrix H of G is a $R^{|V|*|E|}$ matrix that can be defined as follows

$$h(v,e) = \begin{cases} 1 \ if \ vertex \ v \ belongs \ to \ hyperedge \ e \\ 0 \ otherwise \end{cases}$$

From the above definition, we can define the degree of vertex v and the degree of hyper-edge e as follows

$$d(v) = \sum_{e \in E} w(e) * h(v,e)$$

$$d(e) = \sum_{v \in V} h(v,e)$$

Let $D_v$ and $D_e$ be two diagonal matrices containing the degrees of vertices and the degrees of hyper-edges in their diagonal entries respectively. Please note that $D_v$ is the $R^{|v|*|v|}$ matrix and $D_e$ is the $R^{|e|*|e|}$ matrix.

## 2.2 Definition of the un-normalized hypergraph Laplacian

The un-normalized hypergraph Laplacian is defined as follows

$$L = D_v - HWD_e^{-1}H^T$$

## 2.3 Properties of L

1. For every vector $f \in R^{|V|}$, we have

$$f^T L f = \frac{1}{2} \sum_{e \in E} \sum_{\{u,v\} \subseteq E} \frac{w(e)}{d(e)} (f(u) - f(v))^2$$

2. L is symmetric and positive-definite

3. The smallest eigenvalue of L is 0, the corresponding eigenvector is the constant one vector 1

4. L has $|V|$ non-negative, real-valued eigenvalues $0 \leq \lambda_1 \leq \lambda_2 \leq \cdots \leq \lambda_{|V|}$

Proof:

1. We know that

$$\frac{1}{2} \sum_{e \in E} \sum_{\{u,v\} \subseteq E} \frac{w(e)}{d(e)} (f(u) - f(v))^2$$

$$= \frac{1}{2} \sum_{e \in E} \sum_{\{u,v\} \subseteq E} \frac{w(e)}{d(e)} (f(u)^2 + f(v)^2 - 2f(u)f(v))$$

$$= \sum_{e \in E} \sum_{u,v \in V} \frac{w(e)}{d(e)} (f(u)^2 - f(u)f(v)) h(u,e) h(v,e)$$

$$= \sum_{e \in E} \sum_{u \in V} w(e) f(u)^2 h(u,e) \sum_{v \in V} \frac{h(v,e)}{d(e)} - \sum_{e \in E} \sum_{u,v \in V} \frac{w(e)}{d(e)} f(u) f(v) h(u,e) h(v,e)$$

$$= \sum_{e \in E} \sum_{u \in V} w(e) f(u)^2 h(u,e) - \sum_{e \in E} \sum_{u,v \in V} \frac{w(e)}{d(e)} f(u) f(v) h(u,e) h(v,e)$$

$$= \sum_{u \in V} f(u)^2 \sum_{e \in E} w(e) h(u,e) - \sum_{e \in E} \sum_{u,v \in V} \frac{w(e)}{d(e)} f(u) f(v) h(u,e) h(v,e)$$

$$= \sum_{u \in V} f(u)^2 d(u) - \sum_{e \in E} \sum_{u,v \in V} \frac{w(e)}{d(e)} f(u) f(v) h(u,e) h(v,e)$$

$$= f^T D_v f - f^T H W D_e^{-1} H^T f$$

$$= f^T (D_v - HW D_e^{-1} H^T) f$$

$$= f^T L f$$

2. L is symmetric follows directly from its own definition.

Since for every vector $f \in R^{|V|}$, $f^T L f = \frac{1}{2} \sum_{e \in E} \sum_{\{u,v\} \subseteq E} \frac{w(e)}{d(e)} (f(u) - f(v))^2 \geq 0$. We conclude that L is positive-definite.

3. The fact that the smallest eigenvalue of L is 0 is obvious.

   Next, we need to prove that its corresponding eigenvector is the constant one vector 1.

   Let $d_v \in R^{|V|}$ be the vector containing the degrees of vertices of hypergraph G, $d_e \in R^{|E|}$ be the vector containing the degrees of hyper-edges of hypergraph G, $w \in R^{|E|}$ be the vector containing the weights of hyper-edges of G, $1 \in R^{|V|}$ be vector of all ones, and $one \in R^{|E|}$ be the vector of all ones. Hence we have

   $$L1 = (D_v - HWD_e^{-1}H^T)1 = d_v - HWD_e^{-1}d_e = d_v - HW one = d_v - Hw = d_v - d_v = 0$$

4. (4) follows directly from (1)-(3).

## 2.4 The definitions of symmetric normalized and random walk hypergraph Laplacians

The symmetric normalized hypergraph Laplacian (defined in [9,10]) is defined as follows

$$L_{sym} = I - D_v^{-\frac{1}{2}} HWD_e^{-1}H^T D_v^{-\frac{1}{2}}$$

The random walk hypergraph Laplacian (defined in [9,10]) is defined as follows

$$L_{rw} = I - D_v^{-1} HWD_e^{-1} H^T$$

## 2.5 Properties of $L_{sym}$ and $L_{rw}$

1. For every vector $f \in R^{|V|}$, we have

$$f^T L_{sym} f = \frac{1}{2}\sum_{e \in E}\sum_{\{u,v\}\subseteq E} \frac{w(e)}{d(e)}\left(\frac{f(u)}{\sqrt{d(u)}} - \frac{f(v)}{\sqrt{d(v)}}\right)^2$$

2. $\lambda$ is an eigenvalue of $L_{rw}$ with eigenvector u if and only if $\lambda$ is an eigenvalue of $L_{sym}$ with eigenvector $w = D_v^{\frac{1}{2}}u$

3. $\lambda$ is an eigenvalue of $L_{rw}$ with eigenvector u if and only if $\lambda$ and u solve the generalized eigen-problem $Lu = \lambda D_v u$

4. 0 is an eigenvalue of $L_{rw}$ with the constant one vector 1 as eigenvector. 0 is an eigenvalue of $L_{sym}$ with eigenvector $D_v^{\frac{1}{2}}1$

5. $L_{sym}$ is symmetric and positive semi-definite and $L_{sym}$ and $L_{rw}$ have $|V|$ non-negative real-valued eigenvalues $0 \leq \lambda_1 \leq \cdots \leq \lambda_{|V|}$

Proof:

1. The complete proof of (1) can be found in [9].
2. (2) can be seen easily by solving

$$L_{sym}w = \lambda w \Leftrightarrow \left(I - D_v^{-\frac{1}{2}}HWD_e^{-1}H^T D_v^{-\frac{1}{2}}\right)w = \lambda w$$

$$\Leftrightarrow D_v^{-\frac{1}{2}}\left(I - D_v^{-\frac{1}{2}}HWD_e^{-1}H^T D_v^{-\frac{1}{2}}\right)w = \lambda D_v^{-\frac{1}{2}}w$$

$$\Leftrightarrow D_v^{-\frac{1}{2}}w - D_v^{-1}HWD_e^{-1}H^T D_v^{-\frac{1}{2}}w = \lambda D_v^{-\frac{1}{2}}w$$

Let $u = D_v^{-\frac{1}{2}}w$, (in the other words, $w = D_v^{\frac{1}{2}}u$), we have

$$L_{sym}w = \lambda w \Leftrightarrow u - D_v^{-1}HWD_e^{-1}H^T u = \lambda u$$

$$\Leftrightarrow (I - D_v^{-1}HWD_e^{-1}H^T)u = \lambda u$$

$$\Leftrightarrow L_{rw}u = \lambda u$$

This completes the proof.

3. (3) can be seen easily by solving

$$L_{rw}u = \lambda u \Leftrightarrow (I - D_v^{-1}HWD_e^{-1}H^T)u = \lambda u$$

$$\Leftrightarrow D_v(I - D_v^{-1}HWD_e^{-1}H^T)u = \lambda D_v u$$

$$\Leftrightarrow (D_v - HWD_e^{-1}H^T)u = \lambda D_v u$$

$$\Leftrightarrow Lu = \lambda D_v u$$

This completes the proof.

4. First, we need to prove that $L_{rw}1 = 0$.

Let $d_v \in R^{|V|}$ be the vector containing the degrees of vertices of hypergraph G, $d_e \in R^{|E|}$ be the vector containing the degrees of hyper-edges of hypergraph G, $w \in R^{|E|}$ be the vector containing the weights of hyper-edges of G, $1 \in R^{|V|}$ be vector of all ones, and $one \in R^{|E|}$ be the vector of all ones. Hence we have

$$L_{rw}1 = (I - D_v^{-1}HWD_e^{-1}H^T)1$$

$$= 1 - D_v^{-1}HWD_e^{-1}d_e$$

$$= 1 - D_v^{-1}HW one$$

$$= 1 - D_v^{-1}Hw$$

$$= 1 - D_v^{-1}d_v$$

$$= 0$$

The second statement is a direct consequence of (2).

5. The statement about $L_{sym}$ is a direct consequence of (1), then the statement about $L_{rw}$ is a direct consequence of (2).

### III. Algorithms

Given a set of proteins $\{x_1, \ldots, x_l, x_{l+1}, \ldots, x_{l+u}\}$ where $n = l + u$ is the total number of proteins (i.e. vertices) in the hypergraph G=(V,E) and given the incidence matrix H of G. The method constructing H from the gene expression data will be described clearly in the Experiments and Results section.

Define c be the total number of functional classes and the matrix $F \in R^{n*c}$ be the estimated label matrix for the set of proteins $\{x_1, \ldots, x_l, x_{l+1}, \ldots, x_{l+u}\}$, where the point $x_i$ is labeled as sign($F_{ij}$) for each functional class j ($1 \leq j \leq c$). Please note that $\{x_1, \ldots, x_l\}$ is the set of all labeled points and $\{x_{l+1}, \ldots, x_{l+u}\}$ is the set of all un-labeled points.

Let $Y \in R^{n*c}$ the initial label matrix for n proteins in the hypergraph G be defined as follows

$$Y_{ij} = \begin{cases} 1 \text{ if } x_i \text{ belongs to functional class } j \text{ and } 1 \leq i \leq l \\ -1 \text{ if } x_i \text{ does not belong to functional class } j \text{ and } 1 \leq i \leq l \\ 0 \text{ if } l+1 \leq i \leq n \end{cases}$$

Our objective is to predict the labels of the un-labeled points $x_{l+1}, \ldots, x_{l+u}$. Basically, all proteins in the same hyper-edge should have the same label.

### Random walk hypergraph Laplacian based semi-supervised learning algorithm

In this section, we will give the brief overview of the random walk hypergraph Laplacian based semi-supervised learning algorithm. The outline of the new version of this algorithm is as follows

1. Construct $D_v$ and $D_e$ from the incidence matrix H of G
2. Construct $S_{rw} = D_v^{-1} H W D_e^{-1} H^T$
3. Iterate until convergence
   $F^{(t+1)} = \alpha S_{rw} F^{(t)} + (1-\alpha)Y$, where $\alpha$ is an arbitrary parameter belongs to [0,1]
4. Let $F^*$ be the limit of the sequence $\{F^{(t)}\}$. For each protein functional class j, label each protein $x_i$ ($l+1 \leq i \leq l+u$) as sign($F_{ij}^*$)

Next, we look for the closed-form solution of the random walk graph Laplacian based semi-supervised learning. In the other words, we need to show that

$$F^* = \lim_{t \to \infty} F^{(t)} = (1-\alpha)(I - \alpha S_{rw})^{-1} Y$$

Suppose $F^{(0)} = Y$. Thus, by induction,

$$F^{(t)} = \alpha^t S_{rw}^t Y + (1-\alpha) \sum_{i=0}^{t-1} (\alpha S_{rw})^i Y$$

Since $S_{rw}$ is the stochastic matrix, its eigenvalues are in [-1,1]. Moreover, since 0<α<1, thus

$$\lim_{t \to \infty} \alpha^t S_{rw}^t = 0$$

$$\lim_{t \to \infty} \sum_{i=0}^{t-1} (\alpha S_{rw})^i = (I - \alpha S_{rw})^{-1}$$

Therefore,

$$F^* = \lim_{t \to \infty} F^{(t)} = (1-\alpha)(I - \alpha S_{rw})^{-1} Y$$

Now, from the above formula, we can compute $F^*$ directly.

**Symmetric normalized hypergraph Laplacian based semi-supervised learning algorithm**

Next, we will give the brief overview of the symmetric normalized hypergraph Laplacian based semi-supervised learning algorithm can be obtained from [9,10]. The outline of this algorithm is as follows

1. Construct $D_v$ and $D_e$ from the incidence matrix H of G
2. Construct $S_{sym} = D_v^{-\frac{1}{2}} H W D_e^{-1} H^T D_v^{-\frac{1}{2}}$
3. Iterate until convergence
   $F^{(t+1)} = \alpha S_{sym} F^{(t)} + (1-\alpha) Y$, where α is an arbitrary parameter belongs to [0,1]
4. Let $F^*$ be the limit of the sequence $\{F^{(t)}\}$. For each protein functional class j, label each protein $x_i$ ($l+1 \le i \le l+u$) as sign($F_{ij}^*$)

Next, we look for the closed-form solution of the normalized graph Laplacian based semi-supervised learning. In the other words, we need to show that

$$F^* = \lim_{t \to \infty} F^{(t)} = (1-\alpha)(I - \alpha S_{sym})^{-1} Y$$

Suppose $F^{(0)} = Y$. Thus, by induction

$$F^{(t)} = \alpha^t S_{sym}^t Y + (1-\alpha) \sum_{i=0}^{t-1} (\alpha S_{sym})^i Y$$

Since $S_{sym}$ is similar to $S_{rw}$ ($S_{rw} = D_v^{-1} H W D_e^{-1} H^T = D_v^{-\frac{1}{2}} S_{sym} D_v^{\frac{1}{2}}$) which is a stochastic matrix, eigenvalues of $S_{sym}$ belong to [-1,1]. Moreover, since 0<α<1, thus

$$\lim_{t \to \infty} \alpha^t S_{sym}^t = 0$$

$$\lim_{t \to \infty} \sum_{i=0}^{t-1} (\alpha S_{sym})^i = (I - \alpha S_{sym})^{-1}$$

Therefore,

$$F^* = \lim_{t \to \infty} F^{(t)} = (1-\alpha)(I - \alpha S_{sym})^{-1} Y$$

Now, from the above formula, we can compute $F^*$ directly.

**Un-normalized hypergraph Laplacian based semi-supervised learning algorithm**

Finally, we will give the brief overview of the un-normalized hypergraph Laplacian based semi-supervised learning algorithm. The outline of this algorithm is as follows

1. Construct $D_v$ and $D_e$ from the incidence matrix H of G
2. Construct $L = D_v - H W D_e^{-1} H^T$
3. Compute closed form solution $F^* = \gamma (L + \gamma I)^{-1} Y$, where γ is any positive parameter
4. For each protein functional class j, label each protein $x_i$ ($l+1 \le i \le l+u$) as sign($F_{ij}^*$)

The closed form solution $F^*$ of un-normalized hypergraph Laplacian based semi-supervised learning algorithm will be derived clearly and completely in Regularization Framework section.

### IV. Regularization Frameworks

In this section, we will develop the regularization framework for the symmetric normalized hypergraph Laplacian based semi-supervised learning iterative version. First, let's consider the error function

$$E(F) = \frac{1}{2}\left\{\sum_{e \in E} \sum_{\{u,v\} \subseteq E} \frac{w(e)}{d(e)} \left\|\frac{F_u}{\sqrt{d(u)}} - \frac{F_v}{\sqrt{d(v)}}\right\|^2\right\} + \gamma \sum_{i=1}^{|V|} \|F_i - Y_i\|^2$$

In this error function $E(F)$, $F_i$ and $Y_i$ belong to $R^c$. Please note that c is the total number of protein functional classes and $\gamma$ is the positive regularization parameters. Hence

$$F = \begin{bmatrix} F_1^T \\ \vdots \\ F_{|V|}^T \end{bmatrix} \text{ and } Y = \begin{bmatrix} Y_1^T \\ \vdots \\ Y_{|V|}^T \end{bmatrix}$$

Here $E(F)$ stands for the sum of the square loss between the estimated label matrix and the initial label matrix and the sum of the changes of a function F over the hyper-edges of the hypergraph [9].

Hence we can rewrite $E(F)$ as follows

$$E(F) = trace(F^T L_{sym} F) + \gamma trace((F - Y)^T (F - Y))$$

Our objective is to minimize this error function. In the other words, we solve

$$\frac{\partial E}{\partial F} = 0$$

This will lead to

$$\left(I - D_v^{-\frac{1}{2}} H W D_e^{-1} H^T D_v^{-\frac{1}{2}}\right) F + \gamma(F - Y) = 0$$

$$F - D_v^{-\frac{1}{2}} H W D_e^{-1} H^T D_v^{-\frac{1}{2}} F + \gamma F = \gamma Y$$

$$F - \frac{1}{1+\gamma} D_v^{-\frac{1}{2}} H W D_e^{-1} H^T D_v^{-\frac{1}{2}} F = \frac{\gamma}{1+\gamma} Y$$

$$\left(I - \frac{1}{1+\gamma} D_v^{-\frac{1}{2}} H W D_e^{-1} H^T D_v^{-\frac{1}{2}}\right) F = \frac{\gamma}{1+\gamma} Y$$

Let $\alpha = \frac{1}{1+\gamma}$. Hence the solution $F^*$ of the above equations is

$$F^* = (1 - \alpha)(I - \alpha D_v^{-\frac{1}{2}} H W D_e^{-1} H^T D_v^{-\frac{1}{2}})^{-1} Y$$

Please note that $S_{rw} = D_v^{-1} H W D_e^{-1} H^T$ is not the symmetric matrix, thus we cannot develop the regularization framework for the random walk hypergraph Laplacian based semi-supervised learning iterative version.

Next, we will develop the regularization framework for the un-normalized hypergraph Laplacian based semi-supervised learning algorithms. First, let's consider the error function

$$E(F) = \frac{1}{2}\left\{\sum_{e \in E}\sum_{\{u,v\} \subseteq E} \frac{w(e)}{d(e)}||F_u - F_v||^2\right\} + \gamma \sum_{i=1}^{|V|}||F_i - Y_i||^2$$

In this error function $E(F)$, $F_i$ and $Y_i$ belong to $R^c$. Please note that c is the total number of protein functional classes and $\gamma$ is the positive regularization parameters. Hence

$$F = \begin{bmatrix} F_1^T \\ \vdots \\ F_{|V|}^T \end{bmatrix} \text{ and } Y = \begin{bmatrix} Y_1^T \\ \vdots \\ Y_{|V|}^T \end{bmatrix}$$

Here $E(F)$ stands for the sum of the square loss between the estimated label matrix and the initial label matrix and the sum of the changes of a function F over the hyper-edges of the hypergraph [9].

Hence we can rewrite $E(F)$ as follows

$$E(F) = F^T LF + \gamma trace((F - Y)^T(F - Y))$$

Please note that un-normalized hypergraph Laplacian matrix is $L = D_v - HWD_e^{-1}H^T$. Our objective is to minimize this error function. In the other words, we solve

$$\frac{\partial E}{\partial F} = 0$$

This will lead to

$$LF + \gamma(F - Y) = 0$$

$$(L + \gamma I)F = \gamma Y$$

Hence the solution $F^*$ of the above equations is

$$F^* = \gamma(L + \gamma I)^{-1}Y$$

Similarly, we can also obtain the other form of solution $F^*$ of the normalized graph Laplacian based semi-supervised learning algorithm as follows (note the symmetric normalized hypergraph Laplacian matrix is $L_{sym} = I - D_v^{-\frac{1}{2}}HWD_e^{-1}H^T D_v^{-\frac{1}{2}}$)

$$F^* = \gamma(L_{sym} + \gamma I)^{-1}Y$$

V.  **Experiments and Results**

In this paper, we use the dataset available from [11] and the references therein. This dataset contains the gene expression data measuring the expression of 4062 S. cerevisiae genes under the set of 215 titration experiments and these proteins are annotated with 138 GO Biological Process functions. In the other words, we are given gene expression data ($R^{4062*215}$) matrix and the annotation (i.e. the label) matrix ($R^{4062*138}$). Every expression values are normalized to z-transformed score such that every gene expression profile has the mean 0 and the standard deviation 1.

Given the gene expression data, we can define the co-expression similarity $s_{ij}$ of gene i and gene j as the absolute value of the Pearson's correlation coefficient between their gene expression profiles. We have $s(i,j) = |corr(g(i,:), g(j,:))|$, where g(i,:) and g(j,:) are gene expression profiles of gene i and gene j respectively. We can define the adjacency matrix A ($R^{4062*4062}$) as follows

$$A(i,j) = \begin{cases} 1 \; if \; s(i,j) > threshold \\ 0 \; if \; s(i,j) \leq threshold \end{cases}$$

In this paper, without bias, we can set threshold=0.5. Then the un-normalized graph Laplacian based semi-supervised learning method can be applied to this adjacency matrix A. The un-normalized graph Laplacian based semi-supervised learning method (i.e. the current state of the art method in network-based methods for protein function prediction) will be served as the baseline method in this paper and its average accuracy performance measure for 138 GO Biology Process functions will be compared with the average accuracy performance measures of thee hypergraph Laplacian based semi-supervised learning methods. The accuracy performance measure Q is given as follows

$$Q = \frac{True\ Positive + True\ Negative}{True\ Positive + True\ Negative + False\ Positive + False\ Negative}$$

Normally, clustering methods offer a natural way to the problem identifying groups of genes that show very similar patterns of expression and tend to have similar functions [8] (i.e. the possible functional modules) in the gene expression data. In this experiment, we use k-mean clustering method (i.e. the most popular "hard" clustering method) since there exists at least one protein that has one GO Biological Process function only. Without bias, if all genes in the gene expression data have at least two GO Biological Process functions, we will use "soft" k-mean clustering method or fuzzy c-means clustering method. Then each cluster can be considered as the hyper-edge of the hypergraph. By using these hyper-edges, we can construct the incidence matrix H of the hypergraph. To make things simple, we can determine the number of cluster of the k-means method as follows

$$number\ of\ cluster = \sqrt{\frac{number\ of\ proteins}{2}}$$

When H is already computed, we can construct the hypergraph G and apply the random walk, symmetric normalized, and un-normalized hypergraph Laplacian based semi-supervised learning to this hypergraph G. Finally, their average accuracy performance measures for all 138 GO Biological Process functions will be computed. These average accuracy performance measures of the three hypergraph Laplacian based methods are given in the following table 1. In these experiments, the parameter alpha is set to 0.85 and $\gamma = 1$.

**Table 1**

|  | Average Accuracy Performance Measures (%) | | | |
| --- | --- | --- | --- | --- |
|  | Graph (un-normalized) | Hypergraph (un-normalized) | Hypergraph (random walk) | Hypergraph (normalized) |
| 138 GO Biological Process functions | 63.99 | **97.95** | **97.95** | **97.95** |

From the above table, we recognized that the average accuracy performance measures for 138 GO Biological Process function of three hypergraph Laplacian based semi-supervised learning are equal. This will be investigated in the future and in the other biological datasets such as protein-protein interaction networks.

Interestingly, the average accuracy performance measures for 138 GO Biological Process of three hypergraph Laplacian based semi-supervised learning methods are much greater than the average accuracy performance measures of graph Laplacian based semi-supervised learning method.

Please note that three-fold cross validation is used to compute the average accuracy performance measures of all four methods used in this paper.

VI.     Conclusions

We have proposed the detailed algorithms and regularization frameworks of the three un-normalized, symmetric normalized, and random walk hypergraph Laplacian based semi-supervised learning methods applying to protein function prediction problem. Experiments show that these three methods greatly perform better than the un-normalized graph Laplacian based semi-supervised learning method since these three methods utilize the complex relationships among proteins (i.e. not pairwise relationship). Moreover, these three methods can not only be used in the classification problem but also the ranking problem. In specific, given a set of genes (i.e. the queries) involved in a specific disease such as leukemia which is my future research, these three methods can be used to find more genes involved in leukemia by ranking genes in the hypergraph constructed from gene expression data. The genes with the highest rank can then be selected and checked by biology experts to see if the extended genes are in fact involved in leukemia. Finally, these selected genes will be used in cancer classification.

Recently, to the best of my knowledge, the un-normalized graph p-Laplacian based semi-supervised learning method have not yet been developed and applied to protein function prediction problem. This method is worth investigated because of its difficult nature and its close connection to partial differential equation on graph field.